# A Method to Judge the Style of Classical Poetry Based on Pre-trained Model

Ziyao Wang[1],
Lanzhou University,
Lanzhou, China,
2126546982@qq.com,

Jiandong Zhang[2],
Shandong University School of Literature,
Shangdong, China,
932662798@qq.com,

Jun Ma[3],
Lanzhou University,
Lanzhou, China,
majun@lzu.edu.cn,

***Abstract*—One of the important topics in the research field of Chinese classical poetry is to analyze the poetic style. By examining the relevant works of previous dynasties, researchers judge a poetic style mostly by their subjective feelings, and refer to the previous evaluations that have become a certain conclusion. Although this judgment method is often effective, there may be some errors. This paper builds the most perfect data set of Chinese classical poetry at present, trains a BART-poem pre -trained model on this data set, and puts forward a generally applicable poetry style judgment method based on this BART-poem model, innovatively introduces in-depth learning into the field of computational stylistics, and provides a new research method for the study of classical poetry. This paper attempts to use this method to solve the problem of poetry style identification in the Tang and Song Dynasties, and takes the poetry schools that are considered to have a relatively clear and consistent poetic style, such as the Hongzheng Qizi and Jiajing Qizi, Jiangxi poetic school and Tongguang poetic school, as the research object, and takes the poems of their representative poets for testing. Experiments show that the judgment results of the tested poetry work made by the model are basically consistent with the conclusions given by critics of previous dynasties, verify some avant-garde judgments of Mr. Qian Zhongshu, and better solve the task of poetry style recognition in the Tang and Song dynasties.**

## I. INTRODUCTION

Because of the long and consistent tradition of Chinese classical poetics, poets often imitate their predecessors. For example, Jiangxi Poetic School in Song Dynasty learned from Du Fu and respected him as the “first ancestor” [1]; Hongzheng Qizi and Jiajing Qizi imitated the poetry of the Tang Dynasty [2]. These poets' poems clearly reveal their learning and inheritance of some predecessors' poetic styles. How to determine whether a poet's poetry is like the style of previous poets (such as Du Fu and Han Yu) or previous times (such as Tang and Song) is a fairly extensive and important topic in the field of Chinese classical poetry research.

The research methods of previous scholars mainly rely on the conclusions of predecessors who have made the history of poetry final, and combine personal reading experience of poetry. This method, which is limited by subjective perception, sometimes leads to different conclusions from different researchers.

This paper attempts to use NLP technology to solve this problem: the essence of poetry style determination is to judge whether a poem can be classified into a certain type of poetry with a specific style. The probability that a poem is predicted to be classified into this style can be regarded as its similarity to this style of poetry. Based on this, this paper first collected a large number of corpus, trained the BART model for tasks related to poetry text, and then transformed the poetry style determination task into a classification task based on the pre-trained model, selected the appropriate training set to fine tune our model, and finally obtained a certain style determination model.

In order to verify the effectiveness of this method, we use this method to train a style judgment model of Tang and Song poetry, which can determine whether a seven character octave is closer to the style of Tang poetry or Song poetry. The model performed well in the test process, and the poems of the representative poets of Hongzheng Qizi and Jiajing Qizi, who imitated Tang poetry, were very similar to Tang poetry. The typical representative of the Song poetry is the poetry of Jiangxi Poetry School and the Tong Guang Ti style poetry of learning from the Song Dynasty has a very low similarity with the Tang poetry, and a very high similarity with the Song poetry. These results of the model are consistent with the findings in the field of classical poetry research in the past, which not only proves that this paper has solved the problem of judging the style of Tang poetry, but also shows that the method of judging the style of poetry proposed in this paper is effective. The main contributions of this paper are:

●Collect a large amount of data on Chinese classical poetry. Before the author of this paper collected the corpus, the most complete public poetry corpus data set collected about 800,000 poems [11], and the corpus established in this paper collected about 1.2 million poems, and extracted the key characters and theme words of each poem, providing a richer data set for researchers.

- Based on the improved poetry corpus, this paper has trained a BART-poem pre-trained model suitable for the tasks related to Chinese classical poetry. Researchers can fine-tune this model to achieve the tasks of sentence completion and poetry generation.
- This paper puts forward a more objective and effective method to judge the style of poetry without subjective factors, which provides a better research tool for researchers in the field of classical poetry.

### A. *Related Work*

The judgment of poetic style in this paper belongs to the research category of computational stylistics, and is one of the most important research directions in the current quantitative analysis of literature.

In previous studies, the model was constructed to represent the poetry and judge its style by extracting character frequency, word frequency, subject words, keywords and other features. For example, Liu *et al.*, [1] established the first system with computer to study style of classical poetry, by word frequency, genre, and rules of level and oblique tones statistics. Li *et al*., [4] divided poetry into Bold Faction and Grace Faction, and classified the style of Song poetries based on word connection. KNN [5], SVM [6] and other algorithms have also been applied to the judgment task of Song poetry style Similar to literature [4], and the accuracy is no less than 90%.

However, the above models can only roughly represent poetry, and the judgment of style can only achieve the accuracy of two categories. With the development of NLP, the texts can be represented more realistically by pre-trained model and larger data sets. Zhao *et al*., [7] used about 800,000 poems based on GPT2-Chinese-poetry model, and realized the continuation of poems with missing sentences, which provided some help for better understanding of incomplete poems. Wang et al., [8] took Si ku Quan Shu as corpus, and realized the tasks of sentence breaking, punctuation adding and named entity recognition in ancient Chinese based on SiKuBERT. Bert-ccpoem model proposed by Sun *et al.* can be used in downstream applications, including poetry retrieval, recommendation and sentiment analysis [9].

The bart model used in this paper is a sequence-to-sequence pre-trained language model with transformer structure proposed by Lewis et al [10]. It is not only as good as other bert models in the task of natural language understanding, but also has better performance in the task of natural language generation.

The structure of this paper is arranged as follows. The second section introduces the data set used in this paper. The third part mainly introduces the algorithms and models used in the experiment. In the fourth section, the judgment result of poetry style is given and the performance of the algorithm is quantitatively described. Finally, the conclusion of this paper and the prospect of future work are given.

## II. DATA SET

In this section, we mainly introduce the construction of data set, including data collection, data processing and data storage.

Data collection: The data set used in this paper is based on about 800,000 classical poems from Qin Dynasty to modern times, which are open-source provided by Werneror [11]. This data set contains four fields, namely “title”, “dynasty”, “author”, and “content”. In addition, many poems of poets in Jin, Yuan, Ming and Qing Dynasties were collected from various poetry websites through web crawlers to make up for the deficiency of data set. At the same time, because some rare poetry materials have no digital resources, they are recorded by manual. It should be noted that some poems contain some unrecognized uncommon words, so replace them with "?". Finally, we get a data set of about 1.2 million poems.

Data processing: The Thulac tool is selected to segment the poems in the data set, and the deactivated words are removed at the same time [12]. The TF-IDF algorithm is used to extract the subject words from the remaining words, the number of which is 1/12 of the text length. Then, the Word2vec algorithm is used to get the keywords of poetry, the number of which is stipulated as 1/10 of the length of the poetry text[13].

Data storage: After the above processing, this paper has constructed the most complete poetry data set currently open, with a total of about 1.2 million poems, including six fields, namely, two fields of “key characters” and “theme words” have been added. Figure 1 shows some common theme words of Tang and Song poems in the form of word cloud.

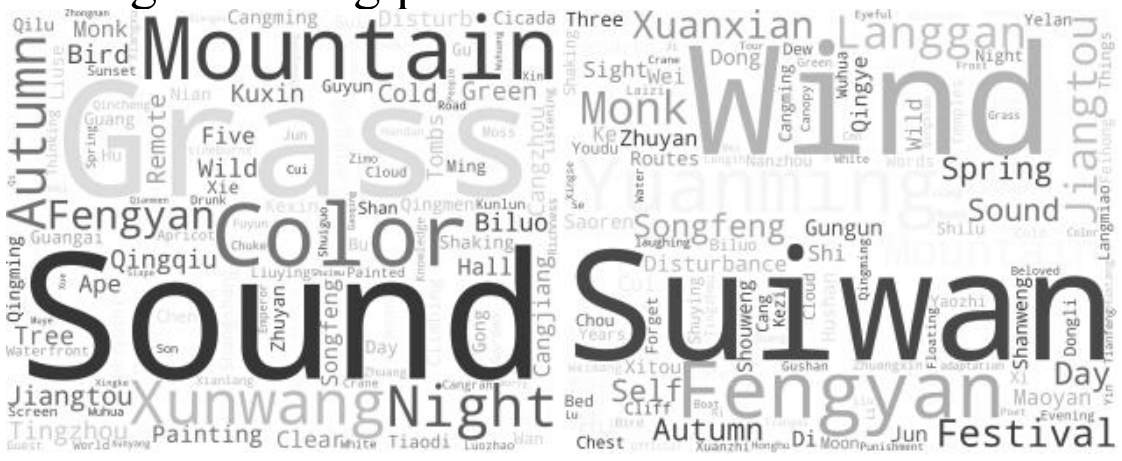

(a) (b)

Figure 1. Theme words of Tang (a) and Song (b) poems

## III. EXPERIMENTAL MODEL AND PROCEDURE

In this chapter, we introduce our model and experimental procedure.

This paper constructs the largest classical poetry dataset at present, and it is necessary to retrain a pre-raining model suitable for poetry tasks based on this dataset. By evaluating the effects of different models, we selected BART for training, and its related parameters are shown in Table 1.

TABLE I. PARAMETERS OF BART MODEL

| parameter | value |
|---|---|
| **Embedding** | 1024 |
| **Feedforward** | 4096 |
| **Hidden** | 1024 |
| **Heads** | 16 |
| **Layer** | 12 |
| **Dropout** | 0.1 |
| **Encoder&decoder** | transformer |

In the pre-trained, data processor is designated as BART mode to pre-process the data [7], and batch_size is set as 64, span_max_length is 3, with 60000 training steps. Finally, the accuracy of the model is stable at 0.91, which is called BART-poem. In addition, it should be noted that, when constructing the data set, "?" is used to replace the unknown words in poetry, and the incomplete text has caused some troubles to the research and judgment. In the process of pre-trained, the BART-poem model can be used to complete the incomplete poems [9], and the experiment proves that good results have been achieved.

Then, through fine-tuning BART-poem model, we can judge the style of poetry. Firstly, the task of judging poetry style is transformed into a multi-category classification problem. Then, two layers of feedforward neural networks are connected to BART-poem model encoder to fine-tune the model. The trained model would return the similarity between the input poetry and each style type, and the model structure is shown in the Figure 2:

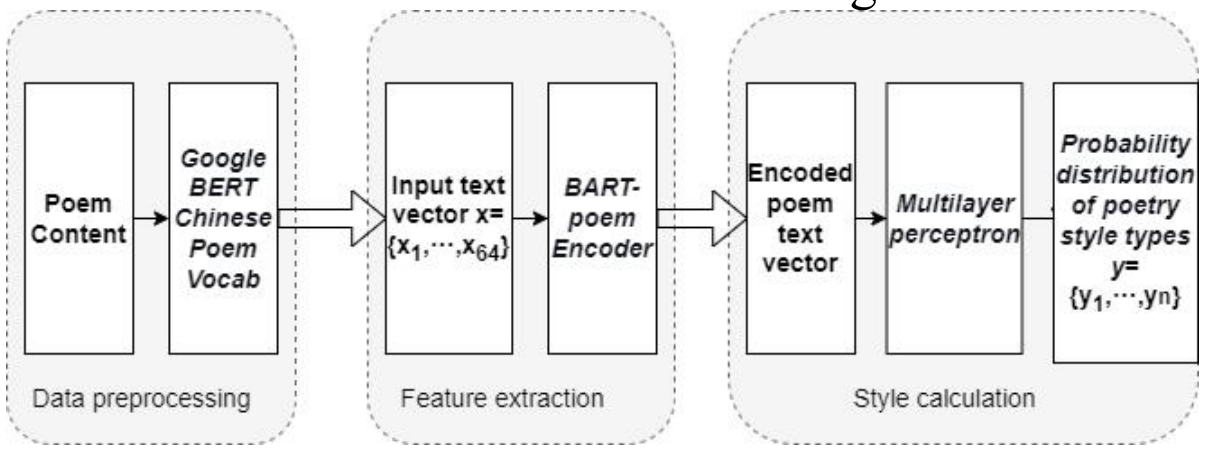

Figure 2. The structure of style judgment model

## IV. EXPERIMENTS

In this part, we introduce our experimental results and experimental analysis.

In this paper, Tang poetry and Song poetry with distinct styles are selected as training data, including about 14,000 Seven-word formal poems, and a model is obtained to judge whether a poem is closer to the style of Tang poetry or Song poetry. The higher the similarity of the model output, the closer the style of the poem is to the corresponding style.

In order to verify the effectiveness of our model, we selected several groups of poems of different poets for experiments. The results are shown in Table 2.

As we all know,There were many poets learning Tang poetry after the Song Dynasty, and the influential ones were the Hongzheng QiZi and the Jiajing QiZi in the Ming Dynasty. It is a consensus in the field of Chinese classical literature that they study Tang poetry in their creation, and the first group of experiments in Table 2 tested the poems of several representative poets of the QiZi.

The experimental results show that their Seven-word formal poems are like Tang poems, and the similarity is about 0.7. However, among several representative poets of QIZI, Wang Shizhen has the lowest similarity with Tang poetry. Qian Zhongshu once proposed that Wang Shizhen learned Song poetry in her later years, which is consistent with the experimental results in this paper.

Jiangxi Poetry School is considered as a typical Song poetry. We selected three representative poets of Jiangxi Poetry School who are not in the training center to test in second group, and found that they have low similarity with Tang poetry and high similarity with Song poetry, which is consistent with the view of the academic community.

In the third group, several representatives of "Tong Guang Ti" Style, a modern poetic school mainly learning from the Song Dynasty, were selected. "Tong Guang Ti" poets mainly studies Song poetry, which is called the Song Poetry School of the Qing Dynasty. The experimental results show that these representative poets of "Tong Guang Ti" Style have low similarity with Tang poetry and high similarity with Song poetry. This experimental result is in line with the fact

In the three groups of experiments, the poetic style judged by our model is basically consistent with its style by manual judgment, which not only proves that this paper has solved the problem of judging the style of Tang poetry, but also shows that the method of judging the style of poetry proposed in this paper is effective.

TABLE II. EXPERIMENTAL RESULTS

| | poets | He JM | Li PL | Xie Z | Wang SZ |
|---|---|---|---|---|---|
| 1 | Tang | 0.70 | 0.71 | 0.70 | 0.52 |
| | Song | 0.30 | 0.29 | 0.30 | 0.48 |
| 2 | poets | Huang TJ | Chen YY | Chen SD | \ |
| | Tang | 0.32 | 0.24 | 0.25 | \ |
| | Song | 0.68 | 0.76 | 0.75 | \ |
| 3 | poets | Chen SL | Zheng XX | Chen BC | Fan DS |
| | Tang | 0.27 | 0.31 | 0.25 | 0.27 |
| | Song | 0.73 | 0.69 | 0.75 | 0.73 |

## V. CONCLUSIONS

Although most poets' works have their own characteristics and originality, it is undeniable that their impart and inheritance is still a very important research factor. The significance of this paper lies in finding a general method that eliminates subjective errors to judge whether a work is like a certain style of poetry, which is an important issue in the field of classical poetry with extensive influence and more discussion.

In order to verify the effectiveness of this method, this paper has established a relatively largest poetry data set at present, including about 1.2 million poems. In order to verify the effectiveness of this method, this paper has established a relatively largest poetry data set at present, including about 1.2 million poems. On this basis, BART model is selected for pre-trained, and poetry generation and completion tasks are performed for the incomplete poetry. After fine-tuning above model, Tang poetry and Song poetry with different styles are selected for testing. The works of three groups of poets in different times are selected for style judgment. The results show that the judgment results of our model are consistent with itself real style. In addition, this study can not only judge the style of classical poetry, but also track and study a poet's learning process, a school of poetry, and even a period of poetic creation.

In the future, we will continue our work, constantly consummate the dataset, and improve our model capability.

## ACKNOWLEDGEMENT

We sincerely thanks to the Kaggle competition platform for polishing the English version of this paper and adjusting some details.